\documentclass[10pt,twocolumn,letterpaper]{article}

\usepackage{cvpr}              

\usepackage{graphicx}
\usepackage{amsmath}
\usepackage{amssymb}
\usepackage{booktabs}
\usepackage{times}
\usepackage{epsfig}
\usepackage{graphicx}
\usepackage{amsmath}
\usepackage{amssymb}
\usepackage{amsthm}
\usepackage{color}
\usepackage{subcaption}
\usepackage{amsmath}

\usepackage{amssymb}
\usepackage{amsmath}
\usepackage{multirow}
\usepackage{newfloat}
\usepackage{listings}
\usepackage{thmtools}
\usepackage{thm-restate}
\usepackage[pagebackref,breaklinks,colorlinks]{hyperref}

\usepackage[capitalize]{cleveref}
\crefname{section}{Sec.}{Secs.}
\Crefname{section}{Section}{Sections}
\Crefname{table}{Table}{Tables}
\crefname{table}{Tab.}{Tabs.}

\def\cvprPaperID{6908} 
\def\confName{CVPR}
\def\confYear{2022}

\begin{document}

\title{LAS-AT: Adversarial Training with Learnable Attack Strategy}

\author{Xiaojun Jia$^{1,2,\dagger}$\thanks{The first two authors contribute equally to this work. $\dagger$ Work done during an internship at Tencent AI Lab. $\ddagger$ Correspondence to: Baoyuan Wu (\href{wubaoyuan@cuhk.edu.cn}{wubaoyuan@cuhk.edu.cn}) and Xiaochun Cao (\href{mailto:caoxiaochun@iie.ac.cn}{caoxiaochun@iie.ac.cn}).} , Yong Zhang$^{3, *}$, Baoyuan Wu$^{4,5,\ddagger}$, Ke Ma$^{6}$, Jue Wang$^{3}$, Xiaochun Cao$^{1,2,\ddagger}$\\
$^{1}$Institute of Information Engineering, Chinese Academy of Sciences, Beijing, China\\
$^{2}$School of Cyberspace Security, University of Chinese Academy of Sciences, Beijing, China\\
$^{3}$ Tencent, AI Lab, Shenzhen, China \\
$^{4}$School of Data Science, The Chinese University of Hong Kong, Shenzhen, China\\
$^{5}$ Secure Computing Lab of Big Data, Shenzhen Research Institute of Big Data, Shenzhen, China \\
$^{6}$ School of Computer Science and Technology, University of Chinese Academy of Sciences, Beijing, China \\
{\tt\small jiaxiaojun@iie.ac.cn; zhangyong201303@gmail.com; wubaoyuan@cuhk.edu.cn;} \\ {\tt\small make@ucas.ac.cn; arphid@gmail.com; caoxiaochun@iie.ac.cn}
}

 
\maketitle


\begin{abstract}
Adversarial training (AT) is always formulated as a minimax problem, of which the performance depends on the inner optimization that involves the generation of adversarial examples (AEs).
Most previous methods adopt Projected Gradient Decent (PGD) with manually specifying attack parameters for AE generation. A combination of the attack parameters can be referred to as an attack strategy. Several works have revealed that using a fixed attack strategy to generate AEs during the whole training phase limits the model robustness and propose to exploit different attack strategies at different training stages to improve robustness.
But those multi-stage hand-crafted attack strategies need much domain expertise, and the robustness improvement is limited.
In this paper, we propose a novel framework for adversarial training by introducing the concept of ``learnable attack strategy", dubbed LAS-AT, which learns to automatically produce attack strategies to improve the model robustness. 
Our framework is composed of a target network that uses AEs for training to improve robustness, and a strategy network that produces attack strategies to control the AE generation. 
Experimental evaluations on three benchmark databases demonstrate the superiority of the proposed method. 
The code is released at \href{https://github.com/jiaxiaojunQAQ/LAS-AT}{https://github.com/jiaxiaojunQAQ/LAS-AT}.
\end{abstract}

\section{Introduction}


Although deep neural networks (DNNs) have achieved great success in academia and industry, they could be easily fooled by adversarial examples (AEs) \cite{szegedy2013intriguing,goodfellow2014explaining}  generated via adding indistinguishable perturbations to benign images. Recently, many studies~\cite{li2019nattack,dong2018boosting,wang2021enhancing,bai2021targeted,bai2020targeted,gu2021adversarial,jia2020adv} focus on generating AEs.
It has been proven that many real-world applications~\cite{kurakin2016adversarial} of DNNs are vulnerable to AEs, such as image classification \cite{goodfellow2014explaining,hirano2021universal}, object detection \cite{wei2018transferable,jia2021effective}, 
neural machine translation \cite{zou2019reinforced,huang2020reinforced}, \emph{etc}. 
The vulnerability of DNNs makes people pay attention to the safety of artificial intelligence and brings new challenges to the application of deep learning~\cite{zheng2020pcal,yin2021adv,xu2020adversarial,gu2021effective,gu2021capsule}. Adversarial training (AT) \cite{madry2017towards,wong2020fast,song2021regional,li2022semi} 
is considered as one of the most effective defense methods to 
improve adversarial robustness by injecting AEs into the training procedure through a minimax formulation. 
Under the minimax framework, the generation of AEs plays a key role in determining robustness. 

\begin{figure}[t]
\begin{subfigure}[t]{0.44\linewidth} 
        \centering
        \includegraphics[width=1\linewidth]{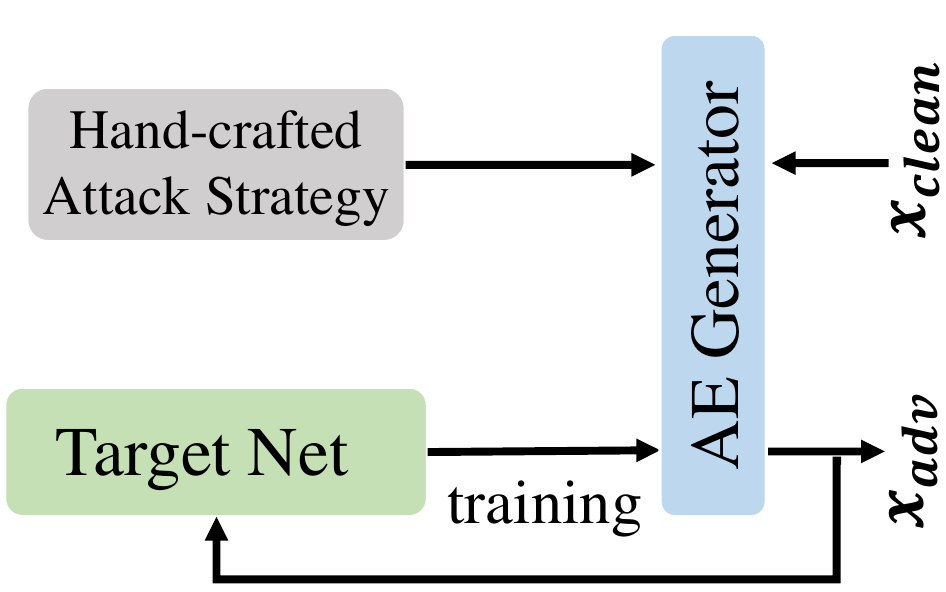}
        \caption{Conventional AT}
        \label{fig:sat0}
    \end{subfigure} 
    \hspace{2mm}
\begin{subfigure}[t]{0.54\linewidth}
        \centering
        \includegraphics[width=1\linewidth]{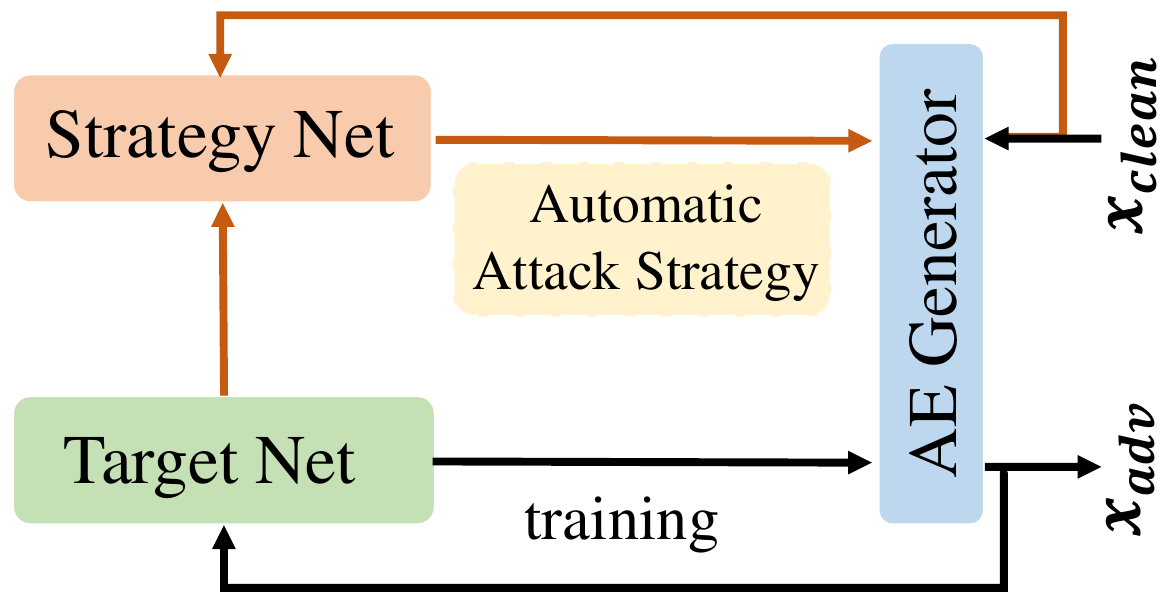}
        \caption{LAS-AT }
        \label{fig:las-at0}
    \end{subfigure} 

\vspace{-3mm}
  \caption{The difference between conventional AT and LAS-AT. (a) Conventional AT methods use a hand-crafted attack strategy to generate AEs. (b) The proposed LAS-AT uses a strategy network to automatically produce sample-dependent attack strategies. 
  }
\label{fig:santan0}
\vspace{-7mm}
\end{figure}

Several recent works improve the standard AT method from different perspectives. Although existing methods~\cite{gowal2020uncovering,dai2021parameterizing,bai2021improving,cui2021learnable,rebuffi2021fixing,jia2021boosting} have made significant progress in improving robustness, they rarely explore the impact of attack strategy on adversarial training.
First, as shown in Fig.~\ref{fig:sat0}, most existing methods leverage a hand-crafted attack strategy to generate AEs by manually specifying the attack parameters, \textit{e.g.,} PGD attack with the maximal perturbation of 8,  iteration of 10, and step size of 2. 
A hand-crafted attack strategy lacks flexibility and might limit the generalization performance. 
Second, most methods use only one attack strategy. Though some works~\cite{cai2018curriculum,wang2019convergence,zhang2020attacks} have realized that exploiting different attack strategies at different training stages could improve robustness, \textit{i.e.,} using weak attacks at the early stages and strong attacks at the late stages, they use manually designed metrics to evaluate the difficulty of AEs and still use one strategy at each stage. However, they need much domain expertise, and the robustness improvement is limited. They use sample-agnostic attack strategies that are hand-crafted and independent of any information of specific samples. There exist statistical differences among samples, and attack strategy should be designed according to the information of the specific sample, \textit{i.e.,} sample-dependent.

\begin{figure}[t]
\begin{center}
   \includegraphics[width=1\linewidth]{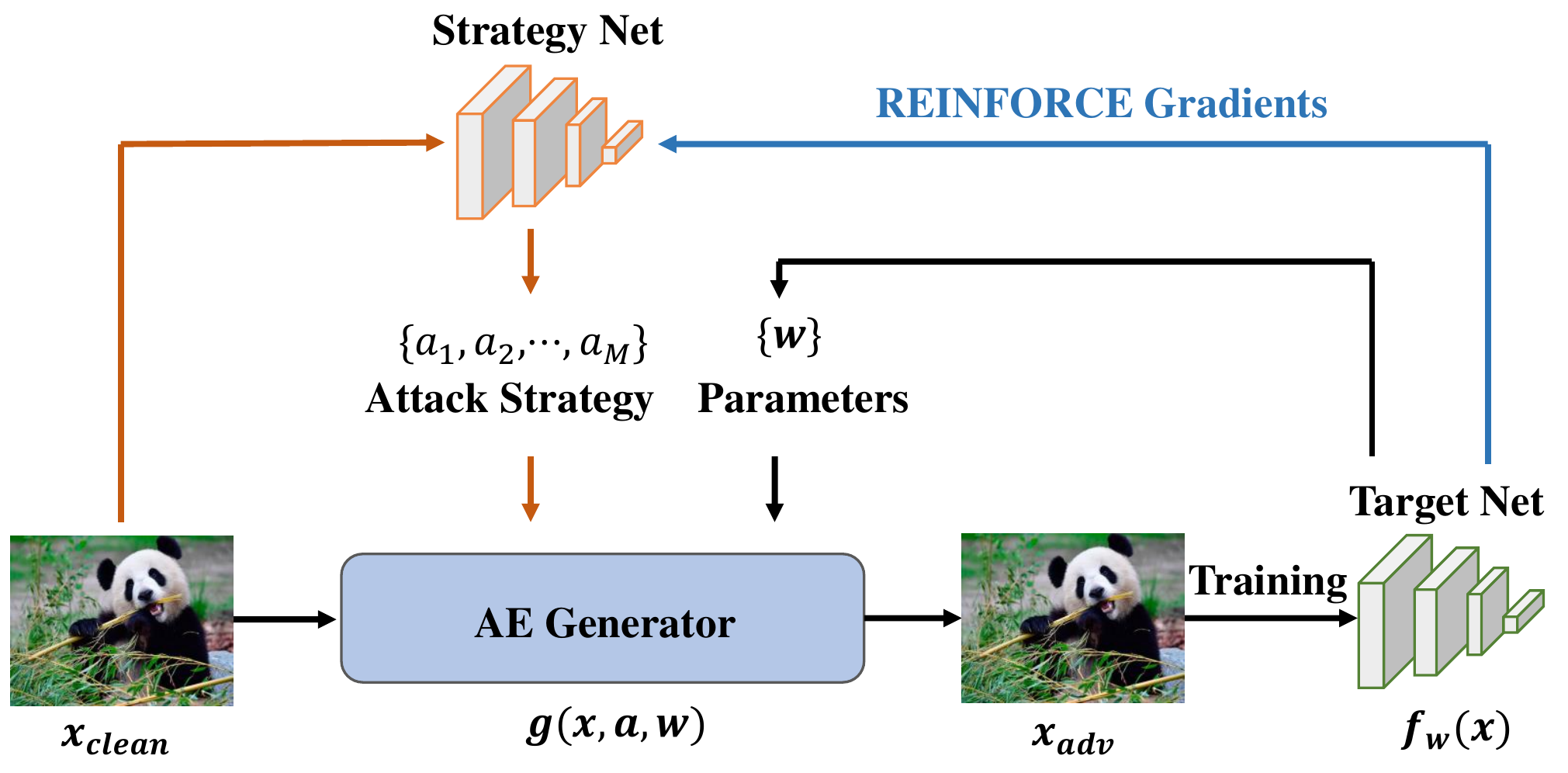}
\end{center}
\vspace{-6mm}
\caption{The framework of proposed LAS-AT. 
It consists of a target network and a strategy network. 
Given a clean image, the strategy network generates an attack strategy. The AE generator takes the strategy as well as the target network to generate an AE which is used to train the target network. 
Some non-differentiable operations (\emph{e.g.} choosing the iteration times) related to attack break gradient flow from the target network to the strategy network. As an alternative approach, REINFORCE algorithm~\cite{williams1992simple} is applied to optimize the strategy network and we utilize the so-called ``REINFORCE gradient" to update the strategy network.
}
\label{fig:santan1}
\vspace{-6mm}
\end{figure}

To alleviate these issues, we propose a novel adversarial training framework by introducing the concept of ``learnable attack strategy", \textit{dubbed LAS-AT}, which learns to automatically produce sample-dependent attack strategies for AE generation  instead of using hand-crafted ones (see Fig.~\ref{fig:las-at0}). 
Our framework consists of two networks, \textit{i.e.,} a target network and a strategy network. 
The former uses AEs for training to improve robustness, while the latter produces attack strategies to control the generation of AEs. 
The two networks play a game where the target network learns to minimize the training loss of AEs while the strategy network learns to generate strategies to maximize the training loss. 
Under such a gaming mechanism, at the early training stages, weak attacks can successfully attack the target network. 
As the robustness improves, the strategy network learns to produce strategies to generate stronger attacks. 
Unlike~\cite{cai2018curriculum} ~\cite{wang2019convergence}, and ~\cite{zhang2020attacks} that use designed metrics and hand-crafted attack strategies, we use the strategy network to automatically produce an attack strategy according to the given sample. 
As the strategy network updates according to the robustness of the target model and the given sample, the strategy network figures out to produce different strategies accordingly at different stages, rather than setting up any manually designed metrics or strategies.  We propose two loss terms to guide the learning of the strategy network.
One evaluates the robustness of the target model updated with the AEs generated by the strategy. 
The other evaluates how well the updated target model performs on clean samples.  
Our main contributions are in three aspects:
\textbf{1)} We propose a novel adversarial training framework by introducing the concept of ``learnable attack strategy", which learns to automatically produce sample-dependent attack strategies to generate AEs. 
    Our framework can be combined with other state-of-the-art methods as a plug-and-play component.
\textbf{2)} We propose two loss terms to guide the learning of the strategy network, which involve explicitly evaluating the robustness of the target model and the accuracy of clean samples.
\textbf{3)} We conduct  experiments and analyses on three databases to demonstrate the effectiveness of the proposed method. 

    

    

\section{Related Work}
\paragraph{Adversarial Attack Methods.}
\label{Related Work}
As the vulnerability of deep learning models has been noticed~\cite{szegedy2013intriguing}, many works studied the model's robustness and proposed a series of adversarial attack methods.  
Fast Gradient Sign Method (FGSM) \cite{goodfellow2014explaining} was a classic adversarial attack method, which made use of the gradient of the model to generate AEs. 
Madry \emph{et al}.~\cite{madry2017towards} proposed a multi-step version of FGSM, called Projected Gradient Descent (PGD). 
To solve the problem of parameter selection in FGSM, Moosavi-Dezfooli \emph{et al}.~\cite{moosavi2016deepfool} proposed a simple but accurate method, called Deepfool, to attack deep neural networks. 
It generated AEs by using an iterative linearization of the classification model. 
Carlini-Wagner \emph{et al}.~\cite{carlini2017towards} proposed several powerful attack methods that could be widely used to evaluate the robustness of deep learning models.  
Moreover, Croce \emph{et al}.~\cite{croce2020reliable} proposed two improved methods (APGD-CE, APGD-DLR) of the PGD-attack. 
They did not need to choose a step size or alternate a loss function. 
And then they combined the proposed method with two complementary adversarial attack methods (FAB \cite{croce2019minimally} and Square \cite{andriushchenko2019square}) to evaluate the robustness, which was called AutoAttack (AA).

\vspace{-4mm}
\paragraph{Adversarial Training Defense Methods.}
Adversarial training is an effective way to improve robustness by using AEs for training, such as ~\cite{ kannan2018adversarial,roth2019adversarial, pang2020boosting, wang2020once, wu2020adversarial, wang2019improving,wang2021probabilistic}. 
The standard adversarial training (AT) is formulated as
a minimax optimization problem in \cite{madry2017towards}.
The objective function is defined as:
\begin{equation}
\min_{\mathbf{w}} \mathbb{E}_{(\mathbf{x}, y) \sim \mathcal{D}}[\max _{\boldsymbol{\delta} \in \Omega} \mathcal{L}(f_{\mathbf{w}}(\mathbf{x} + \boldsymbol{\delta}), y)], \label{Eq:sat}
\end{equation}
where $\mathcal{D}$ represents an underlying data distribution and $\Omega$ represents the perturbation set. 
 $\mathbf{x}$ represents the example, $y$
represents the corresponding label, and $\boldsymbol{\delta}$ represents the indistinguishable perturbation. 
$f_{\mathbf{w}}(\cdot)$ represents the target network and
$\mathcal{L}(f_{\mathbf{w}}(\mathbf{x}), y)$ represents the loss function of the target network.
The inner maximization problem of standard AT can be regarded as the attack strategy that guides the creation of AEs, which is the core to improve the model robustness. A training strategy is designed accordingly, which significantly improves the network's robustness. 
Madry \emph{et al}. proposed the 
prime AT framework, PGD-AT~\cite{
madry2017towards}, to improve the robustness.
And  Rice \emph{et al}. 
proposed a early stopping version~\cite{rice2020overfitting} of PGD-AT, which gained a great improvement. Zhang \emph{et al}.~\cite{zhang2019theoretically} explored a trade-off between standard accuracy and adversarial robustness and proposed a defense method (TRADES) that can trade standard accuracy off against adversarial robustness. Wu \emph{et al.}~\cite{wu2020adversarial}  investigated the weight loss landscape and proposed an effective Adversarial Weight Perturbation (AWP) method to improve the robustness.
Cui \emph{et al}.~\cite{cui2021learnable} proposed to adapt the logits of one model trained on clean data to guide adversarial training (LBGAT). These AT methods adopted a fxed attack strategy to conduct AT. Some AT methods exploited different attack
strategies at different training stages to improve robustness. In detail, 
 Cai \emph{et al}.~\cite{cai2018curriculum}
adopted curriculum adversarial training (CAT) to improve model robustness.
Wang \emph{et al}.~\cite{wang2019convergence} designed a criterion to measure the convergence quality and proposed dynamic adversarial training
(DART) to improve the robustness of the target model. Zhang \emph{et al.}~\cite{zhang2020attacks} proposed to search for the least adversarial data for AT, which could be called friendly adversarial training (FAT). 

\section{The Proposed Approach}
We propose a novel adversarial training framework by introducing the concept of ``learnable attack strategy". 
We first introduce the pipeline of our framework in Sec.~\ref{sec:game} and then present our novel formulation of adversarial training in Sec.~\ref{sec:minmax} and our proposed loss terms in Sec.~\ref{sec:loss} followed by the proposed optimization algorithm in Sec.~\ref{sec:optimize}. 
\subsection{Pipeline of the Proposed Framework} \label{sec:game}
The pipeline of our framework is shown in Fig.~\ref{fig:santan1}. 
Our model is composed of a target network and a strategy network. 
The former uses AEs for training to improve its robustness, whilst the latter generates attack strategies to create AEs to attack the target network. They are competitors. 

\vspace{1mm}
\noindent \textbf{Target Network.} 
The target network is a convolutional network for image classification, denoted as $\hat{y} = f_{\mathbf{w}}(\mathbf{x})$ where $\hat{y}$ is  the estimation of the label, $\mathbf{x}$ is an image, and $\mathbf{w}$ are the parameters of the network.  

\vspace{1mm}
\noindent \textbf{Strategy Network.} 
The strategy network generates adversarial attack strategies to control the AE generation, which takes a sample as input and outputs a strategy. 
Since the strategy network updates gradually, it gives different strategies given the same sample as input according to the robustness of the target network at different training stages. 
The architecture of the strategy network is illustrated in the \textbf{supplementary material}.  Given an image, the strategy network outputs an attack strategy, \textit{i.e.,} the configuration of how to perform the adversarial attack. 
Let $\mathbf{a} = \{a_1, a_2, ..., a_M\} \in \mathcal{A}$ denote a strategy of which each element refers to an attack parameter. 
$\mathcal{A}$ denotes the value space of strategy. 
Parameter $a_m \in \{1,2,...,K_m\}$ has $K_m$ options, which is encoded by a one-hot vector. 
The meaning of each option differs in different attack parameters. 
For example,  PGD attack~\cite{madry2017towards} has three attack parameters, \textit{i.e.,} the attack step size $\alpha$, the attack iteration $I$, and the   maximal perturbation strength $\varepsilon$. 
Each parameter has $K_m$ optional values to select, \textit{e.g.,} the options for $\alpha$ could be $\{0.1, 0.2, 0.3, ...\}$ and the options for $I$ could be $\{1, 10, 20, ...\}$.   
A combination of the selected values for these attack parameters is an attack strategy. 
The strategy is used to created AEs along with the target model. 
The strategy network captures the conditional distribution  of a given $\mathbf{x}$ and $\boldsymbol{\theta}$, $p(\mathbf{a}|\mathbf{x};\boldsymbol{\theta} )$,  where $\mathbf{x}$ is the input image and $\boldsymbol{\theta}$ denotes the strategy network parameters.

\vspace{1mm}
\noindent \textbf{Adversarial Example Generator.} Given a clean image, the process of the generation of AEs can be defined as:
\begin{align}
    \mathbf{x}_{adv}:= \mathbf{x} + \boldsymbol{\delta} \leftarrow  g(\mathbf{x}, \mathbf{a}, \mathbf{w}) , \label{Eq:adv} 
\end{align}
where $\mathbf{x}$ is a clean image, $\mathbf{x}_{adv}$ is its corresponding AEs, and $\boldsymbol{\delta}$ is the generated perturbation. $\mathbf{a}$ is an attack strategy. $\mathbf{w}$ represents the target network parameters, and $g(\cdot)$ is the PGD attack. 
The process is equivalent to solving the inner optimization problem of Eq.~(\ref{Eq:sat}) given an attack strategy $\mathbf{a}$, \textit{i.e.,} finding the optimal perturbation to maximize the loss. 

\subsection{Novel Formulation of Adversarial Training} \label{sec:minmax}
By using Eq.~(\ref{Eq:adv}) that represents the process of AE generation, the standard AT with a fixed attack strategy can be rewritten as: 
\begin{align}
  \min_{\mathbf{w}}  \mathbb{E}_{(\mathbf{x}, y) \sim \mathcal{D}}~  \mathcal{L}(f_{\mathbf{w}}(\mathbf{x}_{adv}), y),  \label{eq:oldAT}
\end{align}
where $\mathbf{x}_{adv} = g(\mathbf{x}, \mathbf{a}, \mathbf{w})$ and $\mathbf{a}$ is the hand-crafted attack strategy. 
$\mathcal{D}$ is the training set. 
$\mathcal{L}$ is the cross-entropy loss function which is used to measure the difference between the predicted label of the AE $\mathbf{x}_{adv}$ and the ground truth $y$. 

Differently, instead of using a hand-crafted sample-agnostic strategy, we use a strategy network to produce automatically generated sample-dependent strategies, \textit{i.e.,} $p(\mathbf{a}|\mathbf{x};\boldsymbol{\theta})$.
Our novel formulation for AT can be defined as: 
\begin{align}
  \min_{\mathbf{w}}  \mathbb{E}_{(\mathbf{x}, y) \sim \mathcal{D}}
  [\max_{\boldsymbol{\theta}}
~\mathbb{E}_{ \mathbf{a} \sim p(\mathbf{a}| \mathbf{x};\boldsymbol{\theta})}~ \mathcal{L}(f_{\mathbf{w}}(\mathbf{x}_{adv}),y) ].
\label{eq:newAT}
\end{align}
Compared to the standard AT, the most distinct difference lies in the generation of AEs, \textit{i.e.,} $\mathbf{x}_{adv}$ \eqref{Eq:adv}. 
The standard AT uses a hand-crafted sample-agnostic strategy $\mathbf{a}$ to solve the inner optimization problem while we use a strategy network to produce the sample-dependent strategy by $p(\mathbf{a}|\mathbf{x};\boldsymbol{\theta})$, \textit{i.e.,} our strategy is learnable. 
Our AE generation involves the parameters $\boldsymbol{\theta}$ of the strategy network, which leads that our loss being a function of the parameters of both networks.  

Comparing Eq.~(\ref{eq:oldAT}) and Eq.~(\ref{eq:newAT}), our formulation is a minimax problem and the inner optimization involves the parameters of the strategy network. 
From Eq.~(\ref{eq:newAT}), it can be observed that the two networks compete with each other in minimizing or maximizing the same objective. 
The target network learns to adjust its parameters to defend AEs generated by the attack strategies, while the strategy network learns to improve attack strategies according to the given samples to attack the target network.
At the beginning of the training phase, the target network is vulnerable, which a weak attack can fool. 
Hence, the strategy network can easily generate effective attack strategies. The strategies could be diverse because both weak and strong attacks can succeed. 
As the training process goes on, the target network becomes more robust. 
The strategy network has to learn to generate attack strategies that create stronger AEs. 
Therefore, the gaming mechanism could boost the robustness of the target network gradually along with the improvement of the strategy network. 

\subsection{The Proposed Loss Terms}
\label{sec:loss}
\vspace{1mm}
\noindent\textbf{Loss of Evaluating Robustness.}
To guide the learning of the strategy network, we propose a new metric to evaluate attack strategy by using the robustness of the one-step updated version of the target model.
Specifically, an attack strategy $\mathbf{a}$ is first used to create an AE $\mathbf{x}_{adv}$ which is then used to adjust the parameters of the target model $\mathbf{w}$ for one step through the first-order gradient descent. 
The attack strategy is criticized to be effective if the updated target model $\hat{\mathbf{w}}$ can correctly predict labels for AEs $\mathbf{x}_{adv}^{\hat{\mathbf{a}}}$ that generated by another attack strategy $\hat{\mathbf{a}}$, \textit{e.g.,} PGD with the maximal perturbation strength of 8, iterative steps of $10$ and step size of $2$. 
The loss function of evaluating robustness can be defined as: 
\begin{align}
   \mathcal{L}_{2}(\boldsymbol{\theta}) 
&=-\mathcal{L}(f( \mathbf{x}_{adv}^{\hat{\mathbf{a}}},\hat{\mathbf{w}} ), y), \label{eq:loss_robust}
\end{align}
where $\hat{\mathbf{w}} = \mathbf{w} - \lambda \nabla_\mathbf{w} \mathcal{L}_1|_{\mathbf{x}_{adv}}$ is the parameters of the updated target network and $\lambda$ is the step size. 
$\mathcal{L}_1$ refers to the loss in Eq.~(\ref{eq:newAT}), \textit{i.e.,} $\mathcal{L}_1(\mathbf{w},\boldsymbol{\theta}):= \mathcal{L}(f(\mathbf{x}_{adv}, \mathbf{w}), y)$. 
$\mathbf{x}_{adv}$
is created by the attack strategy $\mathbf{a}$, which is to be evaluated. 
$\mathbf{x}_{adv}^{\hat{\mathbf{a}}} :=  g(\mathbf{x}, \hat{\mathbf{a}}, \hat{\mathbf{w}}) $ is the AE created by another attack strategy $\hat{\mathbf{a}}$, which is used to evaluate the robustness of the updated model $\hat{\mathbf{w}}$. 
Please note that $\mathcal{L}_2$ is used to evaluate the attack strategy and $\mathbf{w}$ is treated as a variable here rather than parameters to optimize.  
Hence, the value of $\mathbf{w}$ is used in Eq.~(\ref{eq:loss_robust}), but the gradient of $\mathcal{L}_2$ will not be backpropagated to update $\mathbf{w}$ through $\hat{\mathbf{w}}$. 
Eq.~(\ref{eq:loss_robust}) indicates that a larger $\mathcal{L}_2$ means the updated target model is more robust, \textit{i.e.,} a better attack strategy. 

\vspace{1mm}
\noindent\textbf{Loss of Predicting Clean Samples.}
A good attack strategy should not only improve the robustness of the target model but also maintain the performance of predicting clean samples, \textit{i.e.,} clean accuracy. 
To further provide guidance for learning the strategy network, we also consider the performance of the one-step updated target model in predicting clean samples. 
The loss of evaluating the attack strategy can be defined as: 
\begin{align}
    \mathcal{L}_3(\boldsymbol{\theta}) &= - \mathcal{L}(f(\mathbf{x}, 
    \hat{\mathbf{w}}), y),  \label{eq:clean_acc}
\end{align}
where $\hat{\mathbf{w}}$ is the same as that in Eq.~(\ref{eq:loss_robust}), \textit{i.e.,} the parameters of the one-step updated target model. 
$\mathcal{L}_3$ is the function of $\boldsymbol{\theta}$ as the AE $\mathbf{x}_{adv}$ involves computing $\hat{\mathbf{w}}$ and $\mathbf{a}$ is the output of the strategy network. 
Eq.~(\ref{eq:clean_acc}) indicates that a larger $\mathcal{L}_3$ means the updated target model has a lower loss in clean samples, \textit{i.e.,} a better attack strategy. 

\paragraph{Formal Formulation.} 
Incorporating the two proposed loss terms, our formulation for AT can be defined as: 
\begin{equation}
        \begin{aligned}
              & & &\min_{\mathbf{w}} \mathbb{E}_{(\mathbf{x}, y) \sim \mathcal{D}}\left[\max_{\boldsymbol{\theta}}~\mathbb{E}_{ \mathbf{a} \sim p(\mathbf{a}| \mathbf{x};\boldsymbol{\theta})}~[\mathcal{L}_1(\mathbf{w},\boldsymbol{\theta})+\right.\\ 
              & & & \left.\textcolor{white}{\min_{\mathbf{w}} \mathbb{E}_{(\mathbf{x}, y) \sim \mathcal{D}}[\max_{\boldsymbol{\theta}}EEEE}\alpha \mathcal{L}_2(\boldsymbol{\theta}) + \beta \mathcal{L}_3(\boldsymbol{\theta})]\right],
        \end{aligned}
        \label{eq:newAT_pro}
    \end{equation}
where $\mathcal{L}_1$ is a function of the parameters of both the target network and the strategy network while $\mathcal{L}_2$ and $\mathcal{L}_3$ involve the parameters of the strategy network. $\alpha$ and $\beta$ are the trade-off hyper-parameters of the two loss terms.

\subsection{Optimization} \label{sec:optimize}
We propose an algorithm to alternatively optimize the parameters of the two networks. 
Given $\boldsymbol{\theta}$, the subproblem of optimizing the target network can be defined as,
\begin{align}
   \min_{\mathbf{w}} \mathbb{E}_{(\mathbf{x}, y) \sim \mathcal{D}}
\mathbb{E}_{ \mathbf{a} \sim p(\mathbf{a}| \mathbf{x};\boldsymbol{\theta})}
[\mathcal{L}_1(\mathbf{w},\boldsymbol{\theta})]. \label{Eq:subprob1}
\end{align}
Given a clean image, the strategy network generates a strategy distribution $p(\mathbf{a}| \mathbf{x};\boldsymbol{\theta})$, and we randomly sample a strategy from the conditional distribution. 
The sampled strategy is used to generate AEs.  
After collecting the AEs for a batch of samples, we can update the parameters of the target model through gradient descent, \textit{i.e.,}
\begin{align}
    \label{eq:update_w}
    \mathbf{w}^{t+1}=\mathbf{w}^{t}-\eta_1 \frac{1}{ N}  \sum_{n=1}^{N} \nabla_{\mathbf{w}}
\mathcal{L}\left(f(\mathbf{x}_{adv}^n, \mathbf{w}^t), y_{n}\right),  
\end{align}
where $N$ is the number of samples in a mini batch and $\eta_1 $ is the learning rate. 

Given $\mathbf{w}$, the subproblem of optimizing the parameters of the strategy network can be written as,
\begin{align}
    \max_{\boldsymbol{\theta}} J(\boldsymbol{\theta}) , 
\end{align}
where $J(\boldsymbol{\theta}) := \mathbb{E}_{(\mathbf{x}, y) \sim \mathcal{D}}
~\mathbb{E}_{ \mathbf{a} \sim p(\mathbf{a}| \mathbf{x};\boldsymbol{\theta})}~ [\mathcal{L}_1 + \alpha \mathcal{L}_2 + \beta \mathcal{L}_3]$. 
The biggest challenge of this optimization problem is that the process of AE generation (see Eq.~(\ref{Eq:adv})) is not differentiable, namely, the gradient can not be backpropagated to the attack strategy through the AEs. 
Moreover, there are some non-differentiable operations (\emph{e.g.} choosing the iteration times) related to attack \cite{peng2018jointly,ilyas2019adversarial}, which sets an obstacle to backpropagate the gradient to the strategy network. 

Following by the REINFORCE algorithm \cite{williams1992simple}, we can compute the derivative of the objective function $J(\boldsymbol{\theta})$ with respect to the parameters $\boldsymbol{\theta}$ as:
\begin{align}
    \nabla_{\boldsymbol{\theta}} J(\boldsymbol{\theta})&= \nabla_{\boldsymbol{\theta}} \mathbb{E}_{(\mathbf{x}, y) \sim \mathcal{D}}
\mathbb{E}_{ \mathbf{a} \sim p(\mathbf{a}| \mathbf{x};\boldsymbol{\theta})}
\left[ \mathcal{L}_{0}\right]  \\
& =  \mathbb{E}_{(\mathbf{x}, y) \sim \mathcal{D}} \int_{\mathbf{a}} \mathcal{L}_{0} \cdot \nabla_{\boldsymbol{\theta}}  p(\mathbf{a}| \mathbf{x};\boldsymbol{\theta}) d\mathbf{a} \nonumber \\
& =  \mathbb{E}_{(\mathbf{x}, y) \sim \mathcal{D}} \int_{\mathbf{a}} \mathcal{L}_{0} \cdot   
p(\mathbf{a}| \mathbf{x};\boldsymbol{\theta}) \nabla_{\boldsymbol{\theta}} \log p(\mathbf{a}| \mathbf{x};\boldsymbol{\theta}) d\mathbf{a} \nonumber \\
& = \mathbb{E}_{(\mathbf{x}, y) \sim \mathcal{D}}
\mathbb{E}_{ \mathbf{a} \sim p(\mathbf{a}| \mathbf{x};\boldsymbol{\theta})} [  \mathcal{L}_{0} \cdot \nabla_{\boldsymbol{\theta}} \log  p(\mathbf{a}| \mathbf{x};\boldsymbol{\theta}) ] \nonumber ,
\end{align}
where $  \mathcal{L}_{0} =  \mathcal{L}_{1} + \alpha  \mathcal{L}_{2} + \beta  \mathcal{L}_{3}$. 
Similar to solving Eq.~(\ref{Eq:subprob1}), we sample attack strategy from the conditional distribution of strategy to generate AEs. 
The gradient with respect to the parameters
can be approximately computed as: 
\begin{align}
    \nabla_{\boldsymbol{\theta}}J(\boldsymbol{\theta}) \approx   \frac{1}{N} \sum_{n=1}^{N} \mathcal{L}_{0}(\mathbf{x}^n;\boldsymbol{\theta}) \cdot \nabla_{\boldsymbol{\theta}} \log p_{\boldsymbol{\theta}} (\mathbf{a}^n|\mathbf{x}^n).
\end{align}
Then, the parameters of the strategy network can be updated through gradient ascent, \textit{i.e.,} 
\begin{align}
    \label{eq:update_theta}
    \boldsymbol{\theta}^{t+1} = \boldsymbol{\theta}^t + \eta_2 \nabla_{\boldsymbol{\theta}}J(\boldsymbol{\theta}^t), 
\end{align}
where $\eta_2$ is the learning rate. And $\boldsymbol{\theta}$ and $\mathbf{w}$ are updated iteratively. We update $\mathbf{w}$ every k times of updating $\boldsymbol{\theta}$.

\subsection{Convergence Analysis}
Based on \eqref{eq:update_w} and \eqref{eq:update_theta}, we have the following convergence result of the proposed adversarial training algorithm.
\vspace{-1mm}
\begin{restatable}{theorem}{convergence}
\label{thm:1}
Suppose that the objective function $\mathcal{L}_0=\mathcal{L}_1+\alpha \mathcal{L}_2+\beta \mathcal{L}_3$ in \eqref{eq:newAT_pro} satisfied the gradient Lipschitz conditions \textit{w.r.t.} $\boldsymbol{\theta}$ and $\mathbf{w}$, and $\mathcal{L}_0$ is $\mu$-strongly concave in $\boldsymbol{\Theta}$, the feasible set of $\boldsymbol{\theta}$.
If $\mathbf{\hat{x}}_{adv}(\mathbf{x},\mathbf{w})$ is a $\sigma$-approximate solution of the $\ell_{\infty}$ ball with radius $\epsilon$ constraint, the variance of the stochastic gradient is bounded by a constant $\sigma^2>0$, and we set the learning rate of $\mathbf{w}$ as 
\begin{equation}
    \eta_1 = \min\left(\frac{1}{L_0},\ \sqrt{\frac{\mathcal{L}_0(\mathbf{w}^0)-\underset{\mathbf{w}}{\min}\ \mathcal{L}_0(\mathbf{w})}{\sigma^2TL_0}}\right),
\end{equation}
where $L_0=L_{\mathbf{w}\boldsymbol{\theta}}L_{\boldsymbol{\theta}\mathbf{w}}/\mu+L_{\mathbf{w}\mathbf{w}}$ is the Lipschitz constants of $\mathcal{L}_0$, it holds that
\begin{equation}
    \frac{1}{T}\sum_{t=0}^{T-1}\mathbb{E}\big[\|\nabla \mathcal{L}_0(\mathbf{w}^t)\|^2_2\big]\leq4\sigma\sqrt{\frac{\Delta L_0}{T}}+\frac{5\delta L^2_{\mathbf{w}\boldsymbol{\theta}}}{\mu},
\end{equation}
where $T$ is the maximum adversarial training epoch number and $\Delta=\mathcal{L}_0(\mathbf{w}^0)-\underset{\mathbf{w}}{\min}\ \mathcal{L}_0(\mathbf{w})$.
\end{restatable}
The detailed proof is presented in the \textbf{supplementary material}. By Theorem 1, if the inner maximization process can obtain a $\delta$-approximation of $\mathbf{x}^*_{\text{adv}}$, the proposed method \textbf{LAS-AT} can archive a stationary point by  sub-linear rate with the precision $5\delta L^2_{\mathbf{w}\boldsymbol{\theta}}/\mu$. Moreover, if $5\delta L^2_{\mathbf{w}\boldsymbol{\theta}}/\mu$ is sufficient small, our method can find the desired robust model $\mathbf{w}^T$ with a good approximation of $\mathbf{x}^*_{\text{adv}}$.

\section{ Experiments}
To evaluate the proposed method, we conduct experiments on three databases, \textit{i.e.,} CIFAR10 \cite{krizhevsky2009learning}, CIFAR100 \cite{krizhevsky2009learning}, and Tiny ImageNet \cite{deng2009ImageNet}. The details of these databases are presented in the \textbf{supplementary material}.

\begin{table}[t]
\centering

\caption{Test robustness (\%) on the CIFAR-10 database using ResNet18. Number in bold  indicates the best. }
\vspace{-3mm}
 \label{table:ablation_k}
  \scalebox{0.8}{
\begin{tabular}{c|c|c|c|c|c|c}
\toprule
Method    & PGD-AT~\cite{rice2020overfitting} & k=1            & k=10  & k=20  & k=40           & k=60  \\ \midrule \midrule
Clean     & 82.56  & \textbf{82.88} & 82.38 & 82.00 & 82.3           & 82.10 \\ \midrule
PGD-10    & 53.15  & 53.71          & 53.89 & 53.53 & \textbf{54.29} & 53.85 \\ \midrule
Time(min) & 261    & 1378           & 432   & 418   & 365            & 333   \\ \bottomrule
\end{tabular}
}
\vspace{-3mm}
\end{table}

\begin{table*}[t]

\centering

\caption{ Test robustness (\%)  on the CIFAR-10 database using WRN34-10.  Number in bold indicates the best. 
 }

\label{tb:cifar10}

\vspace{-3mm}
 \scalebox{1}{
\begin{tabular}{c|cccccc}
\toprule
Method           & Clean          & PGD-10         & PGD-20         & PGD-50         & C\&W           & AA             \\ \midrule \midrule
PGD-AT ~\cite{rice2020overfitting}         & 85.17          & 56.07          & 55.08          & 54.88          & 53.91          & 51.69          \\ 
TRADES ~\cite{zhang2019theoretically}          & 85.72          & 56.75          & 56.1           & 55.9           & 53.87          & 53.40          \\ 
MART ~\cite{wang2019improving}            & 84.17          & 58.98          & 58.56          & 58.06          & 54.58          & 51.10          \\ 
FAT  ~\cite{zhang2020attacks}            & \textbf{87.97} & 50.31          & 49.86          & 48.79          & 48.65          & 47.48          \\ 
GAIRAT ~\cite{zhang2020geometry}           & 86.30          & 60.64          & 59.54          & 58.74          & 45.57          & 40.30          \\ 
AWP  ~\cite{wu2020adversarial}            & 85.57          & 58.92          & 58.13          & 57.92          & 56.03          & 53.90          \\
LBGAT ~\cite{cui2021learnable}           & 88.22          &          56.25      &        54.66        &           54.3     &          54.29      & 52.23          \\ 
 \midrule \midrule
LAS-AT(ours)     & 86.23          & 57.64          & 56.49          & 56.12          & 55.73          & 53.58          \\
LAS-TRADES(ours) & 85.24          & 58.01          & 57.07          & 56.8           & 55.45          & 54.15          \\ 
LAS-AWP(ours)    & 87.74          & \textbf{61.09} & \textbf{60.16} & \textbf{59.79} & \textbf{58.22} & \textbf{55.52} \\ \bottomrule
\end{tabular}
}
\vspace{-3mm}
\end{table*}

\begin{table*}[t]

\centering

\caption{ 
 Test robustness (\%)  on the CIFAR-100 database using WRN34-10.  Number in bold indicates the best.  
}
 \label{tb:cifar100}
 
 \vspace{-3mm}

\begin{tabular}{c|cccccc}
\toprule
Method           & Clean          & PGD-10         & PGD-20         & PGD-50         & C\&W             & AA             \\ \midrule \midrule
PGD-AT~\cite{rice2020overfitting}           & 60.89          & 32.19          & 31.69          & 31.45          & 30.1           & 27.86          \\ 

TRADES~\cite{zhang2019theoretically}           & 58.61          & 29.20          & 28.66          & 28.56          & 27.05          & 25.94          \\ 
SAT~\cite{sitawarin2021sat}              & 62.82          &     28.1           &       27.17         &  26.76              &    27.32            & 24.57          \\ 
AWP~\cite{wu2020adversarial}              & 60.38          & 34.13          & 33.86          & 33.65          & 31.12          & 28.86          \\ 
LBGAT~\cite{cui2021learnable}            & 60.64          &      35.13          &        34.75        &             34.62   &     30.65           & 29.33          \\ \midrule \midrule
LAS-AT(ours)     & 61.80          & 33.45          & 32.77          & 32.54          & 31.12          & 29.03          \\ 
LAS-TRADES(ours) & 60.62          & 32.99          & 32.53          & 32.39          & 29.51          & 28.12          \\ 
LAS-AWP(ours)    & \textbf{64.89} & \textbf{37.11} & \textbf{36.36} & \textbf{36.13} & \textbf{33.92} & \textbf{30.77} \\ \bottomrule
\end{tabular}

 \vspace{-5mm}
\end{table*}
\subsection{Settings}
\label{Settings}
\vspace{1mm}
\noindent\textbf{Competitive Methods.}
To evaluate the proposed method effectiveness in improving the robustness of a target model, we combine it with several state-of-the-art adversarial training methods and illustrate its performance improvements. 
We choose not only the most popular methods such as the early stopping PGD-AT \cite{rice2020overfitting} and TRADES
\cite{zhang2019theoretically} as base models but also the recently proposed method AWP \cite{wu2020adversarial}.
The combinations of our method and these models are referred to as LAS-PGD-AT, LAS-TRADES, and LAS-AWP, respectively. 
Note that we use the \textbf{same training settings} as the base models \cite{rice2020overfitting, zhang2019theoretically, wu2020adversarial} to train our proposed models, including data splits and training losses. Then we compare the proposed LAS-PGD-AT, LAS-TRADES, and LAS-AWP with the following baselines: (1) PGD-AT~\cite{rice2020overfitting}, (2) TRADES ~\cite{zhang2019theoretically}, (3) SAT~\cite{sitawarin2021sat}, (4) MART~\cite{wang2019improving}, (5) FAT~\cite{zhang2020attacks}, (6) GAIRAT~\cite{zhang2020geometry}, (7) AWP ~\cite{wu2020adversarial} and (8) LBGAT~\cite{cui2021learnable}.
Moreover, we compare our method with CAT \cite{cai2018curriculum},  DART \cite{wang2019convergence} and FAT~\cite{zhang2020attacks}
They use different attack strategies at different training stages to conduct AT. Besides, we also compare our method with other state-of-the-art  hyper-parameter search methods \cite{lin2019online, DBLP:conf/iclr/ZhangWZZ20} to evaluate our method.

\noindent\textbf{Evaluation.}\label{Evaluation}
We choose several adversarial attack methods to attack the trained models, including PGD~\cite{madry2017towards},  C\&W~\cite{carlini2017towards}
and AA \cite{croce2020reliable} which consists of  APGD-CE \cite{croce2020reliable}, APGD-DLR \cite{croce2020reliable}, FAB \cite{croce2019minimally} and Square \cite{andriushchenko2019square}. 
Following the default setting of AT, the max  perturbation strength $\epsilon$ is set to 8 for all attack methods under the $L_{\infty}$. 
The clean accuracy and robust accuracy are used as the evaluation metrics.

\noindent\textbf{Implementation Details.}
On CIFAR-10 and CIFAR-100, we use ResNet18 \cite{he2016deep} or WideResNet34-10(WRN34-10) \cite{zagoruyko2016wide} as the target network. 
On Tiny ImageNet, 
we use PreActResNet18 \cite{he2016identity} as the target model. 
For all experiments, we train the target network for defense baselines, following their original papers.
For the training hyper-parameters of the target network of our method, we use the \textbf{same setting} as the base models \cite{rice2020overfitting, zhang2019theoretically, wu2020adversarial}.
The detailed settings are presented in the
\textbf{supplementary material}. 
For the target network,
we adopt SGD momentum optimizer with a learning rate of 0.1, weight decay of $5 \times 10^{-4}$.
We use ResNet18 as the backbone of the strategy network. 
For the strategy network of our method, we adopt SGD momentum optimizer with a learning rate of $0.001$. The trade-off hyper-parameters $\alpha$ and $\beta$ are set to $2.0$ and $4.0$. 
The range of the maximal perturbation strength is set from $3$ to $15$, the range of the attack step is set from $1$ to $6$, and the range of the attack iteration is set from $3$ to $15$. 
\vspace{-1mm}

\subsection{Hyper-parameter Selection}
The hyper-parameter k controls the alternative update of $\mathbf{w}$ and $\boldsymbol{\theta}$. We update $\mathbf{w}$ every k times of updating $\boldsymbol{\theta}$. It 
not only affects model robustness
but also affects model training efficiency. A hyper-parameter selection experiment with ResNet18 is conducted on CIFAR-10 to select the optimal hyper-parameter k. The results are shown in Table~\ref{table:ablation_k}.
The training time of the proposed LAS-PGD-AT decreases along with the increase of parameter k. The more time the strategy network requires for training, the smaller the k is. When k = 40, the proposed LAS-PGD-AT achieves the best adversarial robustness.
Considering AT efficiency, we set k to 40. The selection of hyper-parameters  $\alpha$ and $\beta$  
 is presented in the \textbf{supplementary material}.

\subsection{Comparisons with Other AT Methods}
Our method is a plug-and-play component that can be combined with other AT methods to boost their robustness.

\begin{table}[t]

\centering

\caption{ Test robustness (\%)  on the Tiny Imagenet database using PreActResNet18. Number in bold indicates the best. 
}
 \label{tb:tiny}
 \vspace{-3mm}
\scalebox{0.8}{
\begin{tabular}{c|cccc}
\toprule 
Method     & Clean          & PGD-50         & C\&W    & AA             \\ \midrule \midrule
PGD-AT~\cite{rice2020overfitting}     & 43.98          & 19.98          &  17.6     & 13.78          \\ 
TRADES~\cite{zhang2019theoretically}     & 39.16          & 15.74          &   12.92    & 12.32          \\ 
AWP  ~\cite{wu2020adversarial}       & 41.48          & 22.51          & 19.02 & 17.34          \\ \midrule \midrule
LAS-AT(ours)     & 44.86          & 22.16          & 18.54 & 16.74          \\  
LAS-TRADES(ours)  & 41.38          & 18.36          & 14.5  & 14.08          \\ 
LAS-AWP(ours)     & \textbf{45.26} & \textbf{23.42} & \textbf{19.88} & \textbf{18.42} \\ \bottomrule
\end{tabular}
}
 \vspace{-4mm}
\end{table}
\begin{table}[t]
\centering

\caption{
Test robustness (\%) on the  CIFAR-10 and CIFAR-100 database. Number in bold indicates the best.
}
\vspace{-3mm}
 \label{tb:Comparison_AT_cifar10}
 \scalebox{0.75}{
\begin{tabular}{c|c|c|c|c}
\toprule 
Database                  & Target network                   & Method        & Clean          & AA             \\ \midrule \midrule
\multirow{2}{*}{CIFAR-10}  &  \multirow{2}{*}{WRN70-16} &      Gowal~\emph{et al}~\cite{gowal2020uncovering}    & 85.29          & 57.20          \\ \cline{3-5} 
                          &                                  & LAS-AWP(ours) & \textbf{85.66} & \textbf{57.61} \\ \midrule
\multirow{2}{*}{CIFAR-100} & \multirow{2}{*}{WRN34-20} & LBGAT~\cite{cui2021learnable}        & 62.55          & 30.20          \\ \cline{3-5} 
                          &                                  & LAS-AWP(ours) & \textbf{67.31} & \textbf{31.92} \\ \bottomrule
\end{tabular}

}
\vspace{-3mm}
\end{table}
\vspace{1mm}
\noindent \textbf{Comparisons on CIFAR-10 and CIFAR-100.}
The results on CIFAR-10 and CIFAR-100 are shown in Table~\ref{tb:cifar10} and Table~\ref{tb:cifar100}. 
Analyses are as follows. 
First, the three proposed models outperform their base models under most attack scenarios.
In a lot of cases, our method not only improves the robustness but also improves the clean accuracy of the base models though there is always a trade-off between accuracy and robustness. 
For example, on CIFAR-10
when using WRN34-10 as the target network, our method improves the clean accuracy of powerful AWP by about $2.2\%$ and also improves the performance of AWP under PGD-10 attack and AA attack by about $2.1\%$ and $1.62\%$, respectively. Moreover, the proposed LAS-AWP achieves the best robustness performance under all attack scenarios.
We attribute the improvements to using automatically generated attack strategies instead of  hand-crafted ones. Second, on CIFAR-100, the proposed LAS-AWP not only achieves the highest accuracy on clean images but also achieves the best robustness performance under all attack scenarios.
In detail, our LAS-AWP outperforms the original AWP $4.5\%$ and $1.9\%$ on the clean accuracy and AA attack accuracy, respectively. Moreover, our LAS-AWP outperforms the powerful LBGAT under all attack scenarios. 
\begin{table}[t]
\centering

\caption{ Test robustness (\%) on the CIFAR-10 database using ResNet18. Number in bold indicates the best.}
\vspace{-3mm}
 \label{table:table0}
  \scalebox{0.85}{
\begin{tabular}{ccc|c|cc}
\toprule 
$\mathcal{L}_1$ & $\mathcal{L}_2$ & $\mathcal{L}_3$ & clean         & PGD-10         & AA             \\  \midrule
\checkmark  &    &    & 81.83         & 53.88          & 49.06          \\  \midrule
\checkmark  & \checkmark  &    & 81.54         & 53.98          & 49.34          \\  \midrule
\checkmark  &    & \checkmark  & 81.90         & 53.89          & 49.20          \\  \midrule
\checkmark  & \checkmark  & \checkmark  & \textbf{82.3} & \textbf{54.29} & \textbf{49.89} \\ \bottomrule
\end{tabular}
}
\vspace{-4mm}
\end{table}

\begin{table}[t]
\centering

\caption{Test robustness (\%) on the CIFAR-10 database using WRN34-10. 
Comparisons with Madry, CAT, DART and FAT. 
The results are reported in \cite{zhang2020attacks}. Number in bold indicates the best.}
\vspace{-3mm}
 \label{tb:CAT_DART}
 \scalebox{0.8}{
\begin{tabular}{c|c|ccc}
\toprule
Method      & Clean & FGSM  & PGD-20 & C\&W   \\ \midrule \midrule
Madry-AT~\cite{madry2017towards}     & 87.3  & 56.1  & 45.8   & 46.8  \\ 
CAT~\cite{wang2019convergence}        & 77.43 & 57.17 & 46.06  & 42.28 \\ 
DART~\cite{wang2019convergence}        & 85.03 & 63.53 & 48.70  & 47.27 \\ 
FAT~\cite{zhang2020attacks} & \textbf{87.97} & 65.94 & 49.86 & 48.65 \\ \midrule

LAS-Madry-AT & 84.95 & \textbf{67.16} & \textbf{55.61}  & \textbf{54.31} \\ \bottomrule
\end{tabular}

}
\vspace{-5mm}
\end{table}

\noindent \textbf{Comparisons on Tiny ImageNet.}
Following \cite{lee2020adversarial}, we use PreActResNet18 \cite{he2016identity} as the target model for evaluation on Tiny ImageNet. 
The results are shown in Table~\ref{tb:tiny}. 
As Tiny ImageNet has more classes than CIFAR-10 and CIFAR-100, the defense of AEs is more challenging. 
Our method improves the  clean and adversarial robustness accuracy of the three base models. 

\vspace{1mm}
\noindent \textbf{Comparisons with state-of-the-art robustness model.}
Auto Attack (AA) is a reliable and strong attack method to evaluate model robustness. It consists of three white-box attacks and a black-box attack. The details is introduced in Sec~\ref{Related Work}. Under their leaderboard results~\footnote{https://github.com/fra31/auto-attack}, on CIFAR-10, Gowal~\emph{et al}.~\cite{gowal2020uncovering} study the impact of hyper-parameters ( such as model weight averaging and model size ) on model robustness and adopt WideResNet70-16 (WRN-70-16) to conduct AT, which ranks the 1st  under AA attack without additional real or synthetic data. We also adopt WRN-70-16 for our method. LAS-AWP can boost the model robustness and achieve higher robustness accuracy.
On CIFAR-100, Cui~\emph{et al}. train WideResNet34-20 (WRN-34-20) for LBGAT and achieves state-of-the-art robustness without additional real or synthetic data.
We also adopt WRN-34-20 for our method. LAS-AWP can also achieve higher robustness accuracy. The result is shown Table~\ref{tb:Comparison_AT_cifar10}.

\subsection{Ablation Study}
In our formulation in Eq.~(\ref{eq:newAT_pro}), besides the loss $\mathcal{L}_1$, we propose two additional loss terms to guide the learning of the strategy network, \textit{i.e.,}  the loss of evaluating robustness $\mathcal{L}_2$ and the loss of predicting clean samples $\mathcal{L}_3$. 
To validate the effectiveness of each element in the objective function, we conduct ablation experiments with ResNet18 on CIFAR-10. 
We train four LAS-PGD-AT models by using $\mathcal{L}_1$, $\mathcal{L}_1 \& \mathcal{L}_2$, $\mathcal{L}_1 \& \mathcal{L}_3$, and $\mathcal{L}_1 \& \mathcal{L}_2 \& \mathcal{L}_3$, respectively.
The trained models are attacked by a set of adversarial attack methods. 
The results are shown in Table \ref{table:table0}.
The classification accuracy is the evaluation metric. 
\textit{Clean} represents using clean images for testing while other attack methods use AEs for testing.

Analyses are summarized as follows. 
First, when incorporating the loss $\mathcal{L}_2$ only, the performance of robustness under all attacks improves while the clean accuracy slightly drops. 
When incorporating the loss $\mathcal{L}_3$ only, the clean accuracy improves, but the performance of robustness under partial attacks slightly drops. 
The results show that $\mathcal{L}_2$ contributes more to improve the robustness and $\mathcal{L}_3$ contributes more to improve the clean accuracy. 
Second, using all losses achieves the best performance in robustness as well as the clean accuracy, which indicates that the two losses are compatible and combining them could remedy 
the side effect of independent use.

\begin{figure}[t]
\begin{center}
   \includegraphics[width=0.82\linewidth]{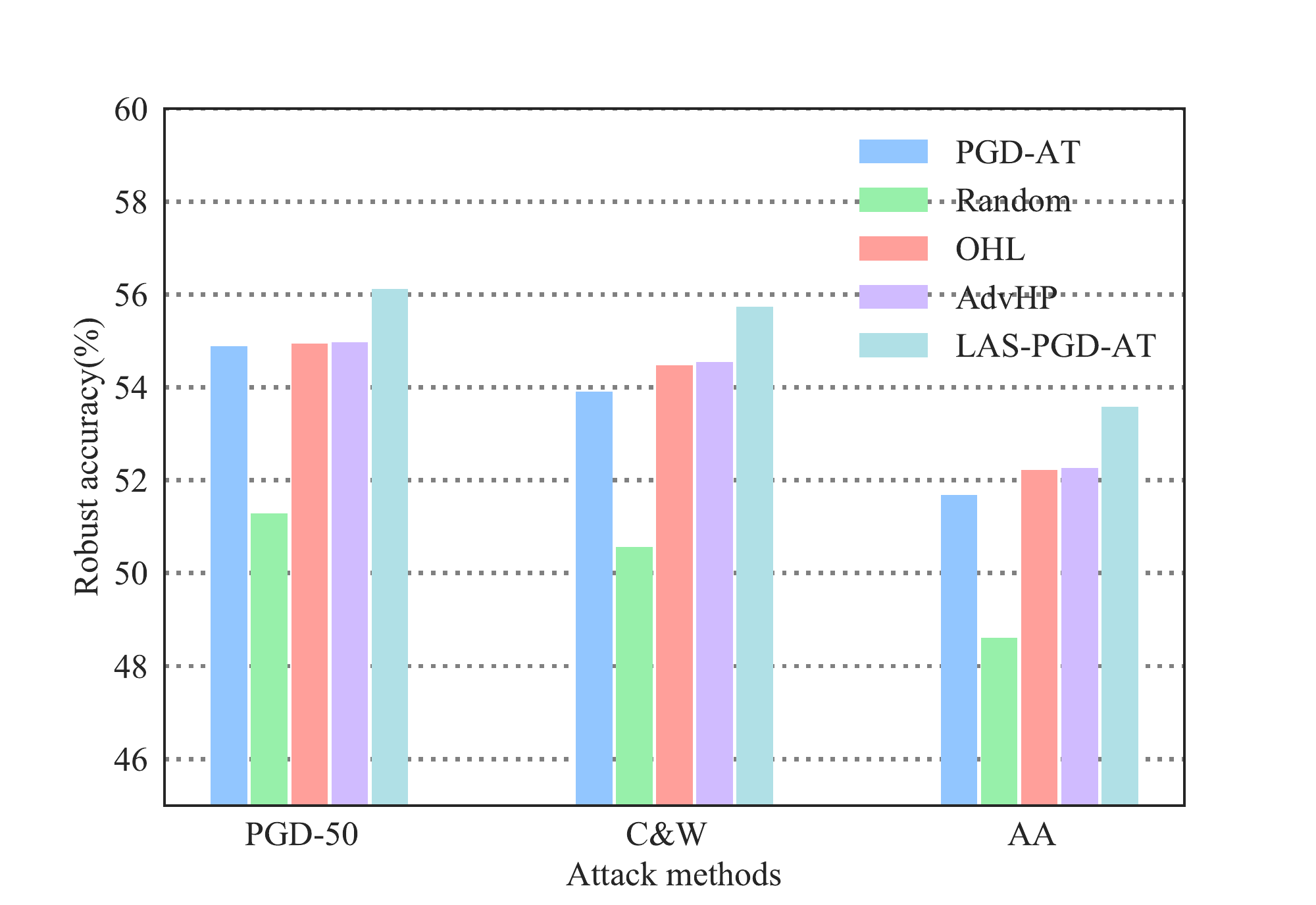}
\end{center}
\vspace{-7mm}
   \caption{Comparisons with the hyper-parameter search  methods  using WRN34-10 on the CIFAR-10 database. $x$-axis represents the attack methods. $y$-axis represents the robust accuracy.  }
\label{fig:comparsion_HP}
\vspace{-5mm}
\end{figure}

\subsection{Performance Analysis}

 \noindent \textbf{Comparisons with hand-crafted attack strategy methods.} 
To investigate the effectiveness of automatically generated attack strategies generated by our method, we compare our LAS-AT with some AT methods (CAT~\cite{cai2018curriculum}, DART~\cite{wang2019convergence} and FAT~\cite{zhang2020attacks}) which adopt dynamic hand-crafted attack strategies for training. 
For a fair comparison, we keep the training and evaluation setting as same as those used in FAT~\cite{zhang2020attacks}. The details are presented in the \textbf{supplementary material}. The result is shown in Table~\ref{tb:CAT_DART}. Our method outperforms competing methods under all attacks. It indicates that compared with previous hand-crafted attack strategies, the proposed automatically generated attack strategies can achieve the greater robustness improvement.

\noindent \textbf{Comparisons with hyper-parameter search methods.
 } 
We compare the proposed method with other hyper-parameter search methods that include a classical hyper-parameter search method (random search) and two automatic  hyper-parameter search methods (OHL~\cite{lin2019online} and AdvHP~\cite{DBLP:conf/iclr/ZhangWZZ20}). For a fair comparison, the same hyper-parameters and search range that are used in our method (see Sec~\ref{Settings}) are adopted for them. The detail settings are presented in the \textbf{supplementary material}. The result is shown in Fig.~\ref{fig:comparsion_HP}. It can be observed that our method achieves the best robustness performance
under all attack scenarios. The automatically generated attack strategies generated by our method are more suitable for AT.

\begin{figure}[t]
\begin{center}
   \includegraphics[width=0.8\linewidth]{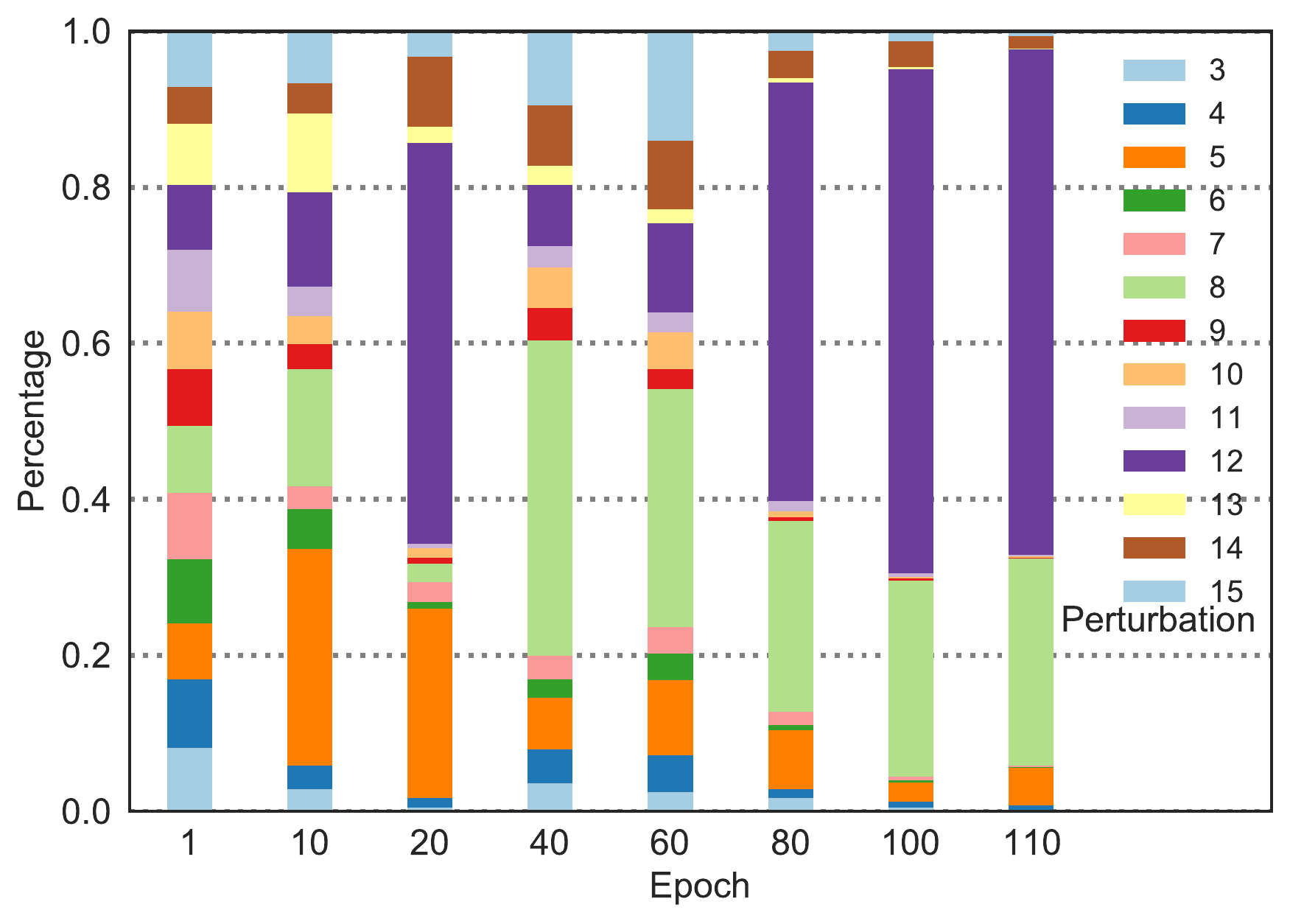}
\end{center}
\vspace{-8mm}
   \caption{The distribution evolution of the maximal perturbation strength in LAS-PGD-AT during training. 
   }
\vspace{-5mm}
\label{fig:discuss}
\end{figure}

\noindent \textbf{Adversarial Training from Easy to Difficult.}
To investigate how LAS-AT works, we analyze the distribution of the strategy network's attack strategies at different training stages.
Experiments using ResNet18 with LAS-PGD-AT are performed on the CIFAR-10 database. 
The range of the maximal perturbation strength is set from 3 to 15. 
The distribution evolution of the maximal perturbation strength during adversarial training is illustrated in Fig.~\ref{fig:discuss}. 

At the beginning of AT, the distribution covers all the optional values of the maximal perturbation strength. 
Each value has a chance to be selected, which ensures the diversity of AEs. 
As the training process goes on, the percentage of small perturbation strengths decreases. 
At the late stages, the distribution of the maximal perturbation strength is occupied by several large values. 
This phenomenon indicates that the strategy network gradually increases the percentage of large perturbation strengths to generate strong AEs because the robustness of the target network is gradually boosted by training with the generated AEs. 
Therefore, it can be observed that under the gaming mechanism, our method starts training with diverse AEs when the target network is vulnerable, and then learns with more strong AEs at the late stages when the robustness of the target network improves.     
 CAT~\cite{wang2019convergence}, DART~\cite{wang2019convergence} and FAT~\cite{zhang2020attacks} adopt hand-crafted strategies to use weak AEs at early stages and then use strong AEs at late stages. 
Unlike them, under our framework, the strategy network automatically generates strategies that determine the difficulty of AEs, according to the robustness of the target network at different stages. 

\section{Conclusion  and Discussion }
We propose a novel adversarial training framework by introducing the concept of ``learnable attack strategy", which is composed of two competitors, \textit{i.e.,} a target network and a strategy network. 
Under the gaming mechanism, the strategy network learns to produce dynamic sample-dependent attack strategies according to the robustness of the target model for adversarial example generation, instead of using hand-crafted attack strategies. 
To guide the learning of the strategy network, we also propose two loss terms that involve evaluating the robustness of the target network and predicting clean samples. 
Extensive experimental evaluations are performed on three benchmark databases to demonstrate the superiority of the proposed method.


\section*{Acknowledgement}
Supported by the National Key R\&D Program of China under Grant 2019YFB1406500, National Natural Science Foundation of China (No.62025604, 62076213, 62006217). Open Project Program of State Key Laboratory of Virtual Reality Technology and Systems, Beihang University (No.VRLAB2021C06). The university development fund of the Chinese University of Hong Kong, Shenzhen under grant No. 01001810, and Tencent AI Lab Rhino-Bird Focused Research Program under grant No.JR202123. 
{\small
\bibliographystyle{ieee_fullname}
\bibliography{egbib}
}
\end{document}